\pdfoutput=1
\documentclass[11pt]{article}

\usepackage[final]{acl}

\usepackage{times}
\usepackage{latexsym}

\usepackage[T1]{fontenc}
\usepackage[utf8]{inputenc}

\usepackage{microtype}

\usepackage{inconsolata}

\usepackage{graphicx,caption}
\usepackage{booktabs}
\usepackage{amsthm,amsmath,amssymb}
\usepackage{mathrsfs}
\usepackage{bm}
\usepackage{multirow,multicol}
\usepackage{newfloat}
\usepackage{listings}
\usepackage{footmisc}
\usepackage{mathabx}

\lstdefinelanguage{json}{
    basicstyle=\small\ttfamily,
    numbers=none,
}

\title{Commentary Generation from Data Records\\ of Multiplayer Strategy Esports Game}

\author{Zihan Wang \\
  The University of Tokyo \\
  \texttt{zwang@tkl.iis.u-tokyo.ac.jp} \\\And
  Naoki Yoshinaga \\
  Institute of Industrial Science, \\
  The University of Tokyo \\
  \texttt{ynaga@iis.u-tokyo.ac.jp} \\}

\begin{document}
\maketitle
\begin{abstract}
Esports, a sports competition on video games, has become one of the most important sporting events. Although esports play logs have been accumulated, only a small portion of them accompany text commentaries for the audience to retrieve and understand the plays. In this study, we therefore introduce the task of generating game commentaries from esports' data records. We first build large-scale esports data-to-text datasets that pair structured data and commentaries from a popular esports game, League of Legends. We then evaluate Transformer-based models to generate game commentaries from structured data records, while examining the impact of the pre-trained language models. Evaluation results on our dataset revealed the challenges of this novel task. We will release our dataset to boost potential research in the data-to-text generation community.\footnote{\url{https://github.com/ArnoZWang/esports-data-to-text}}
\end{abstract}

\section{Introduction}

\begin{figure}[!t]
\centering
\small
\tabcolsep 0pt
\begin{tabular}{p{\linewidth}}
Screenshot of ``\texttt{WARD\_PLACED}'' event (for explanation): \\
\includegraphics[width=\linewidth]{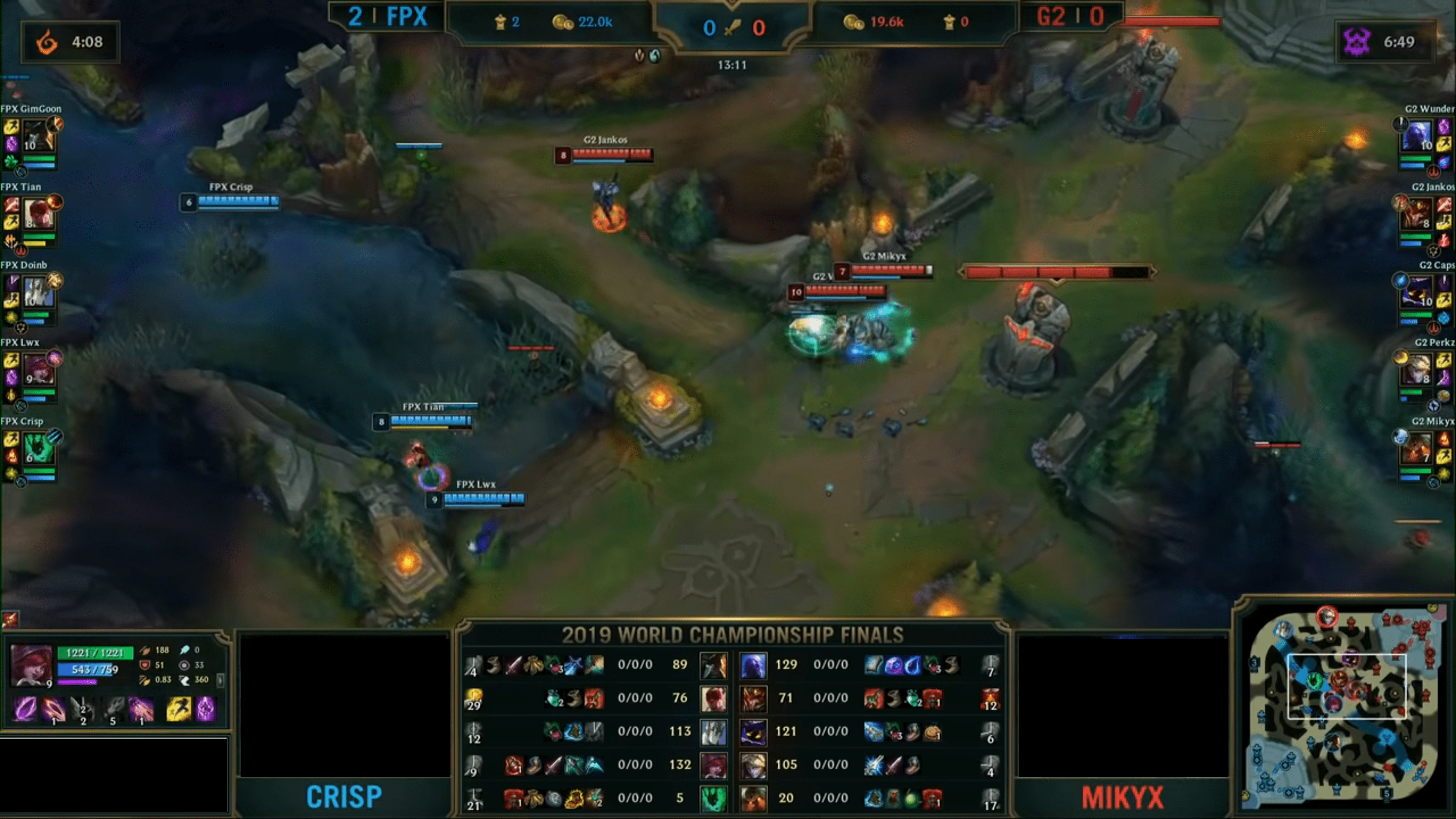} \\
\toprule
\textbf{Input} (one-minute structured data; an excerpt):\\
\begin{minipage}{\linewidth}
\begin{lstlisting}[language=json]
{
 ...
 {"type": "WARD_PLACED", 
  "timestamp": 905433, 
  "wardType": "YELLOWTRINKET", 
  "creatorId": 6
 },
 ...
}
\end{lstlisting}
\end{minipage}
\\
\midrule
\multicolumn{1}{@{}p{0.9\linewidth}}{\textbf{Output} (play-by-play commentary; an excerpt):}\\
\ldots\ even in a map state g2 can get exclusive vision on an area then suddenly the Nautilus veigar will have a lot of zone control but so behind in map control\ \ldots \\
\bottomrule
\end{tabular}
\caption{Game commentary based on one-minute data records (a series of events in JSON format). The screenshot is to help interpret the input (not a part of the input).}
\label{fig:esports_data}
\end{figure}

Esports~\cite{hamari2017esports,reitman2020esports}, a sports competition using video games, has become popular and gained a larger audience than ever. However, the individual gameplays accompany a few metadata such as player names, which prevents their audience from finding games with strategies of interest and understanding the intention of skillful actions. Although textual game commentaries will help the audience retrieve games by a natural language query and
better understand the player's actions (Figure~\ref{fig:esports_data})~\cite{lavelle2010critical}, it is costly for human experts to provide individual games with play-by-play commentaries. As a result, only a small fraction of esports games with play logs accompany textual commentaries.

To enhance the audience's experience in watching such esports games, the technology of data-to-text generation can be used to generate game commentaries from structured data records. In the literature, data-to-text generation has been applied in summary generation from box- and line-scores of basketball games~\cite{wiseman2017challenges} and play-by-play commentary generation for board games~\cite{modgil2013added,kameko2015learning}. Compared to the basketball data, esports data contains detailed game records and play-by-play commentaries. Compared to board game data, esports data is usually not turn-based. In conclusion, there is a lack of research considering the characteristics of esports data in existing studies.

In this study, we introduce the task of generating game commentaries from structured data records (Figure~\ref{fig:esports_data}) for one of the most popular esports games, League of Legends (LoL). Broadly, the overall workflow for addressing this new task includes three main processes: we first build a large-scale data-to-text generation dataset consisting of commentaries obtained from subtitles of YouTube contest videos on esports games and corresponding structured data records obtained using LoL official APIs; we then use a Transformer encoder-decoder model~\cite{vaswani2017attention} and its pre-trained variants to tackle the task; we also set evaluation metrics of esports data-to-commentary generation.

We evaluate the performance of esports data-to-commentary generation on our proposed datasets using the metrics regarding the characteristics of esports data and discuss the main challenges of this task to be addressed in the future.
% We also attempt to address some issues found in the baseline models.

The contributions of this paper are as follows:
\begin{itemize}
\item We set up a task of data-to-commentary generation for one of the most popular esports, League of Legends; we have built large-scale datasets to facilitate the research on this task.
% to facilitate research on the event-specific commentary generation for esports data of multiplayer strategy game.
\item We designed evaluation criteria for esports data-to-commentary generation, which reflects the purposes of the game commentaries.
\item We evaluated several strong baselines including Llama2~\cite{touvron2023llama} and discussed the challenges of the task through examples of generated commentaries.
% \item We analyze the evaluation results and uncover existing challenges of the data-to-commentary generation.
 \end{itemize}

\section{Related Work}
\label{sec:rel}
In this section, we review data-to-text generation tasks on sports games, board games, and video games to highlight the characteristics of our task.

\paragraph{Summary generation for sports game records}
Many studies have focused on generating textual summaries of physical sports games from their score records~\cite{wiseman2017challenges,van2017pass,dou2018data2text,van2018evaluating,taniguchi2019generating,puduppully2019data,rebuffel2020hierarchical}, mainly including basketball and soccer games. The essential difference between these tasks and our task is that their data only records certain important values (score, player number, etc.), while esports data provides details of the games. As a result, the average length of commentaries per game is much longer for our LoL dataset than those for the sports game records, which challenges a generation model to understand individual actions in the games.

\paragraph{Play-by-play commentary generation for board games}
Some studies tackled the task of generating play-by-play commentaries with grounded move expressions from chess and shogi (Japanese chess)~\cite{modgil2013added,kameko2015learning,jhamtani-etal-2018-learning}. For both esports and chess, we can reproduce the whole game from the data records. Nevertheless, in board games, two players alternately perform one move in turn, whereas in esports games, multiple players can simultaneously perform actions in real-time, which challenges a model to interpret simultaneous actions.

\paragraph{Commentary generation from esports game videos}
Several studies aimed to produce textual summaries and commentaries from a game video~\cite{khan2015video,pasunuru2018game,tanaka2021lol,zhang2024game}. In particular, regarding racing games, \citet{ishigaki-etal-2021-generating} generated live commentary from game video data with structured telemetry data, mostly on numerical values of the player's car and game progress. Although this study utilizes game data records, they provide only partial information on the gameplays and are meant to supplement visual data. Meanwhile, our task focuses on League of Legends~\cite{tanaka2021lol}, one of the most popular multiplayer strategic esports games~\cite{zhang-etal-2022-moba}, and generates commentaries from comprehensive structured data records with detailed descriptions of games with more strategic content.

\section{Esports Data-to-text Datasets}
\label{sec:dataset}
We have constructed and will release large-scale data-to-text datasets for one of the most popular esports games, League of Legends (LoL), which is also a demonstration sports event in the 2018 and 2022 Asian Games for its popularity~\cite{hallmann2018esports,jenny2017virtual}. The large-scale dataset is the core building block to assess the feasibility of data-to-text technology on the task. Therefore, in this study, we build two datasets, LoL19 and LoL19-21, from all games in the highest-level tournaments of the 2019 and 2019 to 2021 Season World Championship of League of Legends, respectively.
% We refer to the regular-sized LoL19 dataset and the larger-sized LoL19-21 dataset as ``core'' and ``extended'' datasets.

In this section, we first introduce the basics of the target game, LoL, and then explain methods of collecting data records and textual commentaries of LoL\@. We also explain several data preprocessing methods to improve the collected datasets.

\begin{table*}[t]
\centering
\small
\begin{tabular}{lrrrl}
\toprule
\multirow{2}{*}[-3pt]{\textbf{Event type}} & \multicolumn{2}{c}{\textbf{Proportion}} & \multirow{2}{*}[-3pt]{\textbf{\# keys}} & \multirow{2}{*}[-3pt]{\textbf{Explanation}} \\
\cmidrule(lr){2-3}
& \textbf{LoL19 (core)} & \textbf{LoL19-21 (extended)} & \\
\midrule
\texttt{ITEM\_PURCHASED} &25.1\% & 24.9\% & 3 & The player purchases an item \\
\texttt{ITEM\_SOLD} &1.8\% & 1.8\% &  3 & The player sells an item \\
\texttt{ITEM\_UNDO} &0.6\% & 0.6\% & 4 & The player cancels the purchase of an item \\
\texttt{ITEM\_DESTROYED} &20.2\% & 20.4\% & 3 & The player destroys an item \\
\texttt{BUILDING\_KILL} &1.3\% & 1.3\% & 8 & The team destroys an enemy building \\
\texttt{CHAMPION\_KILL} &2.4\% & 2.5\% & 5 & The player defeats an enemy champion \\
\texttt{SKILL\_LEVEL\_UP} &13.8\% & 13.4\% & 4 & The player upgrades a skill \\
\texttt{WARD\_PLACED} &24.5\% & 24.7\% &  4 & The player places a ward \\
\texttt{WARD\_KILL} &9.8\% & 9.9\% & 3 & The player destroys an enemy ward \\
\texttt{ELITE\_MONSTER\_KILL} &0.6\% & 0.6\% & 4 or 5 & The team defeats an elite monster \\
 \bottomrule
\end{tabular}
\caption{Statistics of all ten types of game events in our LoL datasets.}.
\label{table:event}
\end{table*}

% \begin{table*}[t]
% \centering
% \small
% \begin{tabular}{lccc}
% \toprule
%  & \textbf{\begin{tabular}[c]{@{}c@{}}LoL19 (core)\\(1min segments)\end{tabular}} & \textbf{\begin{tabular}[c]{@{}c@{}}LoL19 (core)\\(30s segments)\end{tabular}} & \textbf{\begin{tabular}[c]{@{}c@{}}LoL19 (core)\\(15s segments)\end{tabular}} \\ \midrule
%  The number of examples (segments) & 3,490 & 6,917 & 13,814 \\ 
%  The average number of events in input & 49.13 & 24.80 & 12.41 \\
%  The average number of tokens of input & 540.47 & 273.20 & 138.12 \\ 
%  The average number of tokens of output & 374.68 & 188.64 & 94.67 \\
% \bottomrule
% \end{tabular}
% \caption{Statistics of the esports data-to-text datasets LoL19 (core) with various data splitting methods.}
% \label{tab:splitting}
% \end{table*}

\subsection{Basics of League of Legends}
In this study, we choose LoL as our research target because of its popularity and representative role as a multiplayer strategy game for esports.
In LoL games, each player controls one game character called ``champion'' with unique abilities that will improve during the game progress and contribute to the team's overall strategy~\cite{cannizzo2015towards}. Two teams compete in one game map, each of which team consists of five players. The goal is to destroy the opponent's base while protecting their own. As the game progresses, the champions can beat the enemy champions, defeat non-player units called ``monsters,'' and destroy buildings to earn resources. They then use the resources to improve their abilities by purchasing items and upgrading skills.

Compared to the physical sports games and board games, LoL is more complicated on real-time game actions, complexity of rules, and data record size per game. These factors are the main obstacles to data-to-text generation.

\subsection{Data Extraction and Preprocessing}
To build the core dataset named LoL19, we target the games of the highest-level tournament in the 2019 Season World Championship of League of Legends. We collect the structured data records of the gameplays from the LoL official API site\footnote{\url{https://developer.riotgames.com/}\label{fn:repeat}} as input and extract subtitles of YouTube videos on the gameplays as output commentaries. This data is strictly paired with the game IDs provided by the game's Match History site.\footnote{\url{https://lol.fandom.com/wiki/2019_Season_World_Championship/Match_History}} Later, we will introduce the method to process the collected data and build large-scale data-to-text dataset.

\subsubsection*{Retrieving Data Records on Gameplays}
In LoL games, every move made by each player is recorded, and the records are available at the LoL official API site.\footref{fn:repeat} 
% including mouse movements and key presses 
From the LoL data, we can strictly restore the entire game from the structured data records, which is impossible in sports games like basketball and soccer. However, the complete data has redundant information, and the large data volume is a heavy burden for storage and subsequent processing.

\begin{table*}[t]
\centering
\small
\tabcolsep 3.5pt
\begin{tabular}{lrrrrr}
\toprule
 & \multicolumn{2}{c}{\textbf{esports data2text}} & \multicolumn{1}{c}{\textbf{esports video2text}} & \multicolumn{1}{c}{\textbf{basketball}} & \multicolumn{1}{c}{\textbf{chess}} \\
 \cmidrule(lr){2-3}
 \cmidrule(lr){4-4}
 \cmidrule(lr){5-5}
 \cmidrule(lr){6-6}
 & \textbf{LoL19 (core)} & \textbf{LoL19-21 (extended)} & \multicolumn{1}{c}{\textbf{LoL-V2T}} & \multicolumn{1}{c}{\textbf{RotoWire}} & \multicolumn{1}{c}{\textbf{GameKnot}} \\ \midrule
 Number of games (matches) & 220 & 650 & 157 & 4,853 & 11,578 \\
 Number of examples & 3,490 & 10,590 & 9,723 & 4,853 & 298,008 \\ 
 Average number of events in input & 49.13 & 48.58 & - & - & - \\
 Average number of tokens of input & 540.47 & 541.10 & - & 628.00 & 25.73 \\ 
 Average number of tokens of output & 374.68 & 373.89 & 15.4 & 337.10 & 20.55 \\
%  \midrule
% The number of event types (Table~\ref{table:event}) & 10 & 10 & N/A & N/A & N/A \\ 
% The average number of events per examples & 49.13 & 48.58 & N/A & N/A & N/A \\
\bottomrule
\end{tabular}
\caption{Statistics of our esports data-to-text datasets and common datasets for similar tasks.}
% , including esports video2text~\cite{tanaka2021lol}, basketball~\cite{wiseman2017challenges}, and chess~\cite{jhamtani-etal-2018-learning}.}
\label{tab:dataset}
\end{table*}

Therefore, we choose another data type provided by the official API, ``event-based data frame.'' In this data, individual gameplays are recorded based on associated events. Each event is defined 
% by the official API 
as an update of certain game status and includes the key named ``\texttt{type},'' which denotes a type of event. Different types of events have various sets of keys; Table~\ref{table:event} lists all event types with their proportions in our LoL dataset. For example, the ``\texttt{WARD\_PLACED}'' events involve information about ``wardType,'' while the ``\texttt{ITEM\_PURCHASED}'' events do not have this key. The event-based data frames 
% of LoL gameplay 
are stored in JSON format, as shown in Figure~\ref{fig:esports_data}.

\subsubsection*{Retrieving Textual Commentaries} We collect YouTube subtitles of the LoL contest videos, which are linked from every contest game in the Match History site, as the output of this task.
The subtitles are split into sentences (precisely, utterances) using line breaks given by YouTube's automatic speech recognition (ASR) as clues. 

Then, we randomly selected 200 examples obtained with the following data formatting and manually confirmed their qualities. The resulting word error rate (WER) was 6.8, which is comparable to human performance on common ASR datasets,\footnote{\url{https://github.com/syhw/wer_are_we}} confirming the data is clean to use for evaluation.

\subsubsection*{Data Formatting} 
The average length of the commentary of one LoL game is over 10K words, which is much longer than the outputs of the existing data-to-text datasets. As a result, we cannot exploit the common Transformer architecture~\cite{vaswani2017attention} for this task. Therefore, we decompose the obtained pairs of structured data and commentaries into pieces of shorter lengths. We first split the sequence of events by the unit duration of one minute of gameplay. We then split the sequence of commentaries by matching their timings with the duration of each subsequence of events.

Next, we address the format difference between the input data (nested list in JSON format) and natural language text to make common encoder-decoder models~\cite{sutskever2014sequence,vinyals2016order} applicable. Specifically, we linearize the structured input. For each key-value pair in the top-level list of each event in the JSON format, we recursively concatenate the value and the key with a delimiter ``\texttt{$\mid$}'' while inserting a space between individual key-value pairs. Following this procedure, ``\texttt{WARD\_PLACED}'' event in the JSON format (Figure~\ref{fig:esports_data}) is linearized into the following sequence:
%\newenvironment{myquote}%
%  {\list{}{\leftmargin=0.2in\rightmargin=0.2in}\item[]}%
%  {\endlist}
%\begin{myquote}
%\fbox{\begin{minipage}{\linewidth}
\begin{quote}
\texttt{WARD\_PLACED$\mid$type 905433$\mid$timestamp YELLOWTRINKET$\mid$wardType 6$\mid$creatorId}
%\end{minipage}}
%\end{myquote}
\end{quote}

% As shown in the example above, we simply add spaces between two adjacent key-value pairs. We also add special tokens (``\texttt{<ent>$\mid$<ent>}'') in the beginning of each event to separate events.

Table~\ref{tab:dataset} lists the statistics of the resulting dataset, LoL19 and other data-to-text datasets (\S~\ref{sec:rel}) such as LoL-V2T~\cite{tanaka2021lol}, Rotowire~\cite{wiseman2017challenges}, and chess~\cite{jhamtani-etal-2018-learning} for comparison. We also collected the data from all games in the 2020 and 2021 Season World Championship of League of Legends to extend the LoL19 dataset (LoL19-21). Our datasets have a comparable number of examples to LoL-V2T and RotoWire and have a comparable number of input and output tokens to RotoWire. To obtain our LoL datasets, we will release the scripts for collecting and processing the data. 

% In the previous data formatting process, we also built the LoL19 (core) dataset in finer unit duration than one minute, specifically, 30 and 15 seconds. The resulting datasets are shown in Table~\ref{tab:splitting}.

% \section{Esports Data-to-text Generation}
% In this section, we xxx. (Merge this section into Experiments?)

\section{Esports Data-to-text Generation}
In this section, we first perform experiments on esports data-to-text generation using the LoL19 dataset and several Transformer~\cite{vaswani2017attention}-based models. Then, we analyze the system outputs to reveal the challenges of this task.

\begin{table*}[t]
\centering
\small
\tabcolsep 2pt
\begin{tabular}{p{425pt}c}
\toprule
\textbf{Strategic depth} & \textbf{score} \\ 
\midrule
Based on the criteria for obtaining a score of 4, the strategic considerations are inspiring, providing insights to help learn from the skillful players and teams & \multirow{2}{*}{5}\\ 
\midrule
Based on the criteria for obtaining a score of 3, the strategic considerations are sufficient and closely related to the game moment described by the structured data & \multirow{2}{*}{4}\\
\midrule
Based on explaining the facts, the commentary also reflects several strategic considerations, such as the player's intention and the team's arrangement & \multirow{2}{*}{3}\\
\midrule
The commentary only reflects the core event of the game moment described by the structured data, without providing any strategic consideration & \multirow{2}{*}{2}\\
\midrule
The commentary reflects no facts or only a few facts of the game moment described by the structured data & 1\\
\bottomrule
\end{tabular}
\caption{Scoring criteria of the strategic depth evaluation.}
\label{table:strategic_depth_score}
\end{table*}

\begin{table*}[t]
\centering
\small
\tabcolsep 3pt
\begin{tabular}{lcccccc}
\toprule
\textbf{Models} & \textbf{\begin{tabular}[c]{@{}c@{}}sacreBLEU\\$\uparrow$\end{tabular}} & \textbf{\begin{tabular}[c]{@{}c@{}}Text distance\\$\downarrow$\end{tabular}} & \textbf{\begin{tabular}[c]{@{}c@{}}ROUGE-L\\$\uparrow$\end{tabular}} & \textbf{\begin{tabular}[c]{@{}c@{}}BERTScore\\$\uparrow$\end{tabular}} & \textbf{\begin{tabular}[c]{@{}c@{}}BARTScore\\$\uparrow$\end{tabular}} & \textbf{\begin{tabular}[c]{@{}c@{}}Strategic depth\\$\uparrow$\end{tabular}} \\
\midrule
Gold & 100 & 0 & 100 & 100 & 0 & 3.164 \\ \midrule
Transformer & 1.4 & 70.22 & 13.62 & 79.06 & -5.27 & 2.312 \\
T5~\cite{raffel2020exploring} & 3.5 & 71.46 & 13.55 & 81.67 & -5.36 & 2.790 \\
Llama2-7B~\cite{touvron2023llama} & 5.1 & 69.01 & 14.98 & 83.16 & -5.02 & 2.994 \\
\textit{\quad w/o finetune} & 0.1 & 98.84 & 0.65 & 63.64 & -5.90 & 2.076 \\
Llama2-13B & \textbf{11.0} & \textbf{63.49} & \textbf{16.94} & \textbf{86.10} & \textbf{-4.61} & \textbf{3.064} \\
\textit{\quad w/o finetune} & 0.2 & 90.03 & 1.00 & 66.56 & -5.86 & 2.170 \\
Llama2-70B (ICL) & 6.2 & 68.82 & 11.93 & 83.54 & -4.77 & 2.916 \\
\bottomrule
\end{tabular}
\caption{Experimental results on the LoL19 esports data-to-text generation dataset.}
\label{table:result_small}
\end{table*}

\subsection{Settings}
\label{sec:settings}
\paragraph{Datasets}
We use the core dataset for evaluation. We first split the games into train, validation, and test sets with a ratio of 8:1:1, according to the chronological order of the individual games, resulting in 2790:350:350 examples. Since the core dataset exclusively comprises data from the 2019 Season World Championship, the terminology used within the games, such as player names, is guaranteed to be consistent across the datasets.

\paragraph{Models}
We compared models based on Transformer~\cite{vaswani2017attention}, T5~\cite{raffel2020exploring}, and variations of Llama2 in the experiments. \textbf{Transformer} is an encoder-decoder model, implemented by OpenNMT\footnote{\url{https://github.com/OpenNMT/OpenNMT-py}} library. \textbf{T5}\footnote{\url{https://huggingface.co/t5-base}} is a pre-trained generative model on text-to-text tasks. \textbf{Llama2} is a 
% one of the most frequently used 
pre-trained large language model
ranging in scale from 7B to 70B parameter~\cite{touvron2023llama}.

\paragraph{Training}
For \textbf{Transformer} and \textbf{T5}, we set decoder dropout of 0.5, training steps of 10,000, and learning rate of 0.001; the other hyperparameters follow their default settings. For \textbf{Llama2-7B} and \textbf{-13B}, we finetune them using QLoRA~\cite{dettmers2024qlora} and 4-bit precision. We set LoRA dropout of 0.1, training steps and learning rate the same with Transformer, and the other hyperparameters follow Huggingface Llama2\footnote{\url{https://huggingface.co/docs/transformers/model_doc/llama2}} document. For \textbf{Llama2-70B}, we apply in-context learning (ICL)~\cite{floridi2020gpt} without updating model weights. The ICL prompt is as follows:
%\begin{myquote}
%\noindent\fbox{\begin{minipage}{\linewidth}
\begin{quote}
\textit{You are an expert of League of Legends esports games. Please read the input data records and describe them in natural language commentary as output. Input: [insert input data here] Output:}
\end{quote}
%\end{minipage}}
%\end{myquote}

\subsection{Evaluation Metrics}
\label{sec:evaluation}
Following the existing data-to-text tasks on sports game summary~\cite{puduppully-etal-2019-data,rebuffel2020hierarchical,tang2023struc}, board game commentary~\cite{jhamtani-etal-2018-learning}, and racing game commentary~\cite{ishigaki-etal-2021-generating}, we adopt \textbf{sacreBLEU}~\cite{papineni2002bleu,post-2018-call}, \textbf{text distance} (normalized Damerau-Levenshtein\footnote{\url{https://github.com/life4/textdistance}})~\cite{brill-moore-2000-improved}, and \textbf{ROUGE-L}~\cite{lin-2004-rouge},\footnote{\url{https://github.com/pltrdy/rouge}} along with \textbf{BERTScore}~\cite{bertscore}\footnote{\url{https://pypi.org/project/bert-score}} and \textbf{BARTScore}~\cite{NEURIPS2021_e4d2b6e6}\footnote{\url{https://github.com/neulab/BARTScore}} for evaluation. These automatic metrics reflect the quality of generated results over correctness and fluency.

Considering the characteristics of multiplayer strategy esports games, assessing the strategic depth of game commentaries is important. The \textbf{strategic depth} is thus designed to measure the extent to which the system output provides useful information on the players' actions. Although the automatic metrics above can arguably measure the general qualities of the system output, we also want the output to contain strategically relevant commentaries, such as reflecting the players' intentions and the team's arrangement regarding the combat.

Because it is difficult to estimate the strategic depth, we gather human scores using criteria tailored for esports commentaries, as detailed in Table~\ref{table:strategic_depth_score}. These scores are collected from five graduate students as human annotators, who understand game rules, the content of contest games, and game commentaries of LoL\@. We calculate the average number of their scores as the results.

\begin{figure}[t]
\centering
\small
\tabcolsep 0pt
\begin{tabular}{p{\linewidth}}
Screenshot of ``\texttt{ITEM\_PURCHASED}'' event (for explanation): \\
\includegraphics[width=\linewidth]{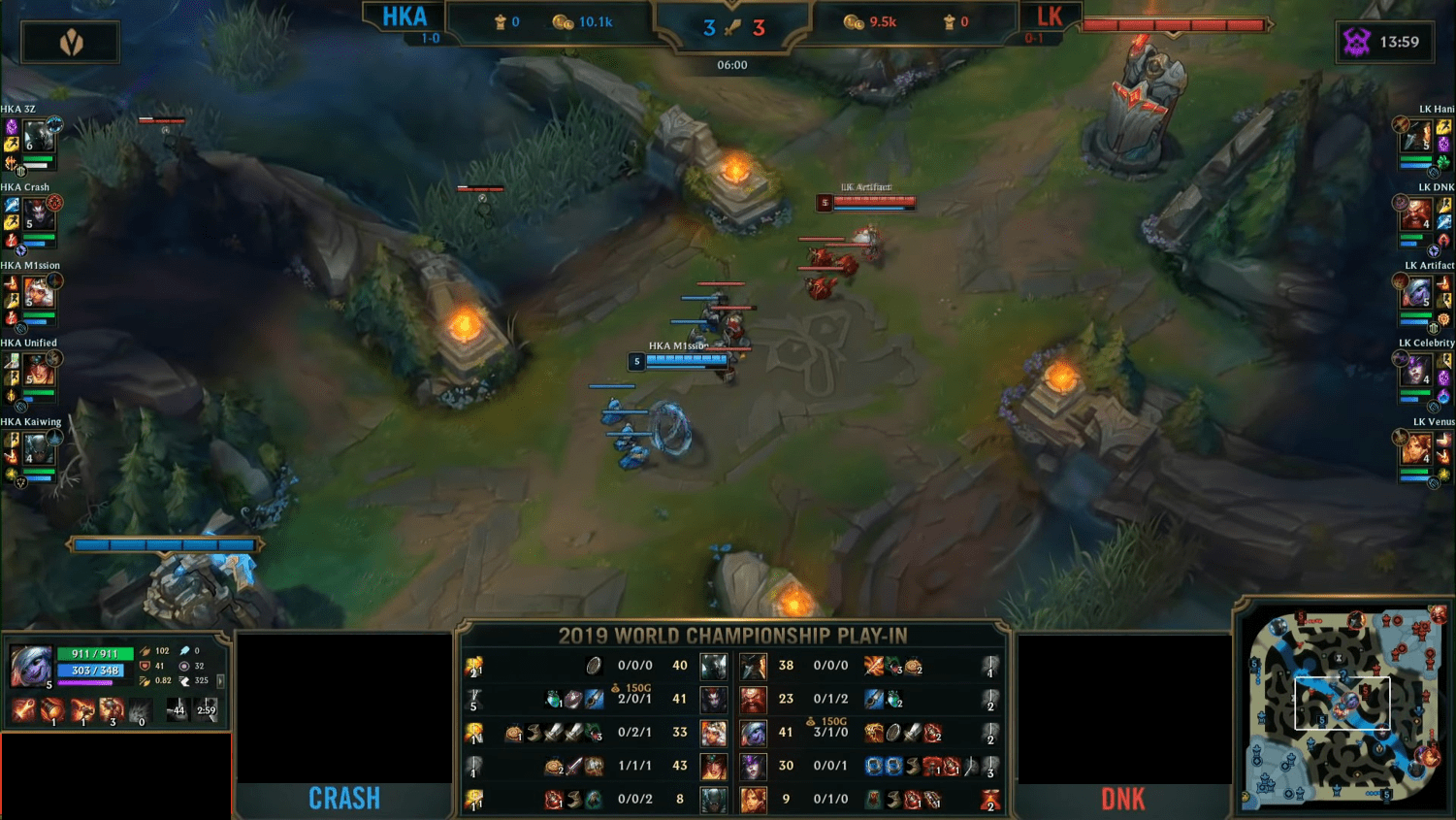} \\
\toprule
\textbf{Input:} \ldots\ \texttt{ITEMPURCHASED|type 783445|timestamp 2|participantId 1033|itemId} \ldots \\
\midrule
\textbf{Reference}: \ldots\ swing Tristana in a very good position I was talking about it yesterday like the build is very forgiving\ \ldots \\
\midrule
\textbf{Transformer}:
\ldots\ I think the rift we are and I think it was slightly changing a little bit more\ \ldots\\
\midrule
\textbf{T5}: \ldots\ that's why you have a look at the fact that 10 minutes into the game\ \ldots\\
\midrule
\textbf{Llama2-7B}: \ldots\ I think that is something that would have been a little bit more active on the map\ \ldots\\
\midrule
\textbf{Llama2-13B}:
\ldots\ they have a lot of pressure in the mid lane which means that they can walk towards the bottom lane of the map if they want to contest\ \ldots\\
\midrule
\textbf{Llama2-70B (ICL)}:
% \ldots\ It seems like there's some interesting activity happening in this game, with items being destroyed and purchased\ \ldots\\
\ldots\ there's some interesting activity\ \ldots\ with items being destroyed and purchased\ \ldots\\
\bottomrule
\end{tabular}
\caption{Excerpts of system outputs of an \texttt{ITEM\_PURCHASED} event. The screenshot is to help interpret the input (not a part of the input).}
\label{fig:visual_example_3}
\end{figure}

\begin{figure}[t]
\centering
\small
\tabcolsep 0pt
\begin{tabular}{p{\linewidth}}
Screenshot of ``\texttt{BUILDING\_KILL}'' event (for explanation): \\
\includegraphics[width=\linewidth]{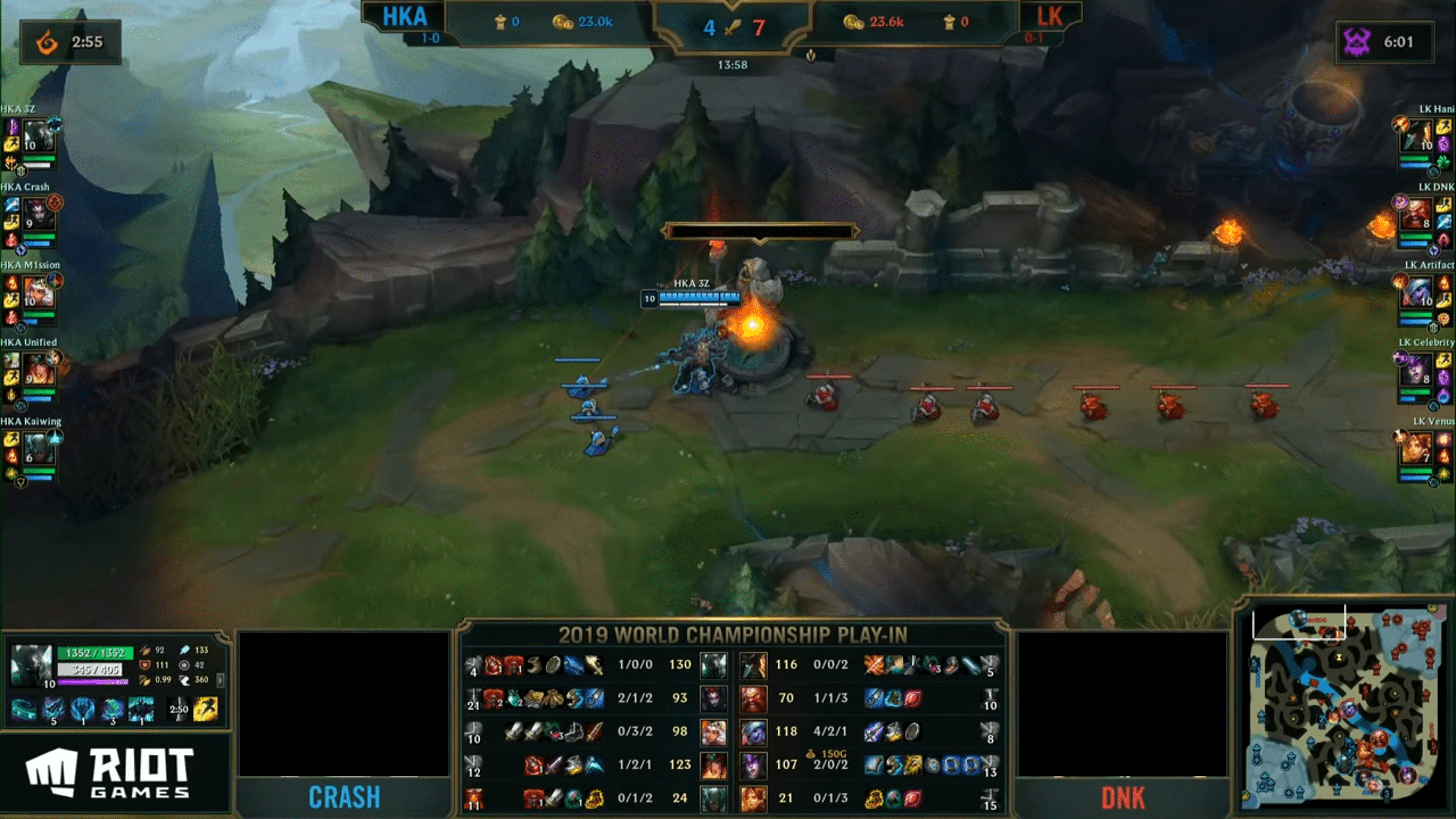} \\
\toprule
\textbf{Input:} \ldots\ \texttt{BUILDINGKILL|type 1315932|timestamp x981y10441|position 8|killerId} \ldots \\
\midrule
\textbf{Reference}: \ldots\ the attacks bit like they can't go anywhere it's always just just like right click the head\ \ldots \\
\midrule
\textbf{Transformer}:
\ldots\ well you can walk away from the bottom Lane you're welcome back\ \ldots\\
\midrule
\textbf{T5}: \ldots\ if you look at the mini-map\ \ldots\ you have to look at the mini-map\ldots\\
\midrule
\textbf{Llama2-7B}: \ldots\ you can see the Kabat does want to be able to get to the side lane\ldots\\
\midrule
\textbf{Llama2-13B}:
\ldots\ at this point in the game you can see that top lane is going to be collected from the side\ \ldots\\
\midrule
\textbf{Llama2-70B (ICL)}:
\ldots\ we do know that the building was a top-lane turret\ \ldots\ and was destroyed by a player\ \ldots\\
\bottomrule
\end{tabular}
\caption{Excerpts of system outputs of an \texttt{BUILDING\_KILL} event. The screenshot is to help interpret the input (not a part of the input).}
\label{fig:visual_example_4}
\end{figure}

\subsection{Results}
\label{sec:experimental_results}
Table~\ref{table:result_small} lists the results on our LoL19 dataset. Compared to the standard Transformer, T5 improves sacreBLEU and strategic depth scores. Llama2-13B exhibits the best overall performance, which confirms the enhancement resulting from pre-trained models with larger sizes. We also compare the performance of Llama2-7B and -13B without finetune as an ablation study. For Llama2-70B, although in-context learning does not result in high text similarity, the generated results exhibit the potential of producing inspiring content.

% Moreover, we analyze the generated results and find remaining challenges of this task. 
Figures~\ref{fig:visual_example_3} and~\ref{fig:visual_example_4} show examples of system outputs, which confirm the difficulty of associating past events with ongoing events. In the first example, the reference output revisits what happened in the past of this game because the ongoing event is not informative enough, while the system outputs mainly focus on only the current moment. In the second example, although the rest of the players are gathered in another area of the map, the lone player in the top lane was taking an enemy building down a little while ago, which directly affected the direction of the whole game. The commentary is expected to be more helpful by reflecting on this moment from the past rather than solely concentrating on the current game moment. In these cases, using finer segments of inputs may enable a more accurate generation. However, it also leads to the loss of context; the system output thus fails to generate content related to the game's history. It is also challenging to maintain the balance between the size of inputs and the amount of context.

Meanwhile, the current focus of this task relies on the modalities of structured data records and textual game commentaries, while visual inputs like screenshots and video clips can provide a contribution to generation. There is a potential to integrate visual inputs for cross-modal generation.

\section{Conclusions}
This study set up the task of generating game commentaries from structured data of the multiplayer strategy game, League of Legends. We built and will release the first large-scale data-to-text generation dataset on strategic esports games. Next, we also discussed evaluation metrics for our task to measure the quality and strategic depth of the system outputs. Then, we explored the performance of Transformer and its pre-trained variants for this task, revealing the challenges of the task such as associating past events with ongoing events.

We will address the remaining issues in the future. The unique challenges include i) linking game history like past events related to current events, and ii) integrating visual inputs, including screenshots and video clips, to this task to perform cross-modal understanding and generation.

\section*{Limitations}
This work mainly focuses on applying data-to-text generation technology in the esports area. Although this paper introduces a novel dataset collected from a representative esports game League of Legends, it lacks the consideration of other esports contests and game genres in the current stage. We plan to continue testing the feasibility of our proposed methods on other esports game data.

\section*{Ethics Statement}
In the data collection process, we have strictly followed the policies of RiotGames API and YouTube. The former is the publisher of LoL game records. The later provides subtitles of LoL contest videos, which we used as game commentaries in our work. Further ethical concerns related to the game content (\textit{e.g.}, video game content rating) can refer to the ESRB Rating (\url{https://www.esrb.org/ratings/32211/league-of-legends/}); the LoL is rated as ``Teen,'' which confirms the game content is suitable ages 13 and up.

\section*{Acknowledgements}
This work was partially supported by JST, CREST Grant Number JPMJCR19A4, Japan and JSPS KAKENHI Grant Number JP21H03494.

% \section*{Acknowledgements}

% This document has been adapted
% by Steven Bethard, Ryan Cotterell and Rui Yan
% from the instructions for earlier ACL and NAACL proceedings, including those for 
% ACL 2019 by Douwe Kiela and Ivan Vuli\'{c},
% NAACL 2019 by Stephanie Lukin and Alla Roskovskaya, 
% ACL 2018 by Shay Cohen, Kevin Gimpel, and Wei Lu, 
% NAACL 2018 by Margaret Mitchell and Stephanie Lukin,
% Bib\TeX{} suggestions for (NA)ACL 2017/2018 from Jason Eisner,
% ACL 2017 by Dan Gildea and Min-Yen Kan, 
% NAACL 2017 by Margaret Mitchell, 
% ACL 2012 by Maggie Li and Michael White, 
% ACL 2010 by Jing-Shin Chang and Philipp Koehn, 
% ACL 2008 by Johanna D. Moore, Simone Teufel, James Allan, and Sadaoki Furui, 
% ACL 2005 by Hwee Tou Ng and Kemal Oflazer, 
% ACL 2002 by Eugene Charniak and Dekang Lin, 
% and earlier ACL and EACL formats written by several people, including
% John Chen, Henry S. Thompson and Donald Walker.
% Additional elements were taken from the formatting instructions of the \emph{International Joint Conference on Artificial Intelligence} and the \emph{Conference on Computer Vision and Pattern Recognition}.

\bibliography{acl_latex,anthology}

\clearpage
% \appendix

% \section{Appendix: In-context Learning of Llama-70B}
% \label{sec:appendix}

% The prompt used for in-context learning of Llama-70B (\S~\ref{sec:settings}) is as follows:
  
% \begin{myquote}
% \fbox{\begin{minipage}{1.04\linewidth}
% \texttt{You are an expert of League of Legends esports games. Please read the input data records and describe them in natural language commentary as output. Input: [insert input data here] Output: }
% \end{minipage}}
% \end{myquote}

\end{document}